\documentclass[11pt]{article}

\usepackage[preprint]{acl}

\usepackage{times}
\usepackage{latexsym}

\usepackage[T1]{fontenc}

\usepackage[utf8]{inputenc}

\usepackage{microtype}

\usepackage{inconsolata}

\usepackage{graphicx}
\usepackage{subfigure}
\usepackage{subcaption}
\usepackage{color}

\usepackage{mdframed}       

\usepackage{amssymb}
\usepackage{amsmath}

\usepackage{multirow}%
\usepackage{amsmath,amssymb,amsfonts}%
\usepackage{amsthm}%
\usepackage{mathrsfs}%
\usepackage[title]{appendix}%
\usepackage{xcolor}%
\usepackage{cite}%
\usepackage{textcomp}%
\usepackage{manyfoot}%
\usepackage{booktabs}%
\usepackage{algorithm}%
\usepackage{algorithmicx}%
\usepackage{algpseudocode}%
\usepackage{listings}%
\usepackage{utfsym}%
\usepackage{tcolorbox}

\definecolor{mainboxbg}{HTML}{F7F9FC}
\definecolor{mainboxborder}{HTML}{A1C6EA}
\definecolor{adminboxbg}{HTML}{FFF4E5}
\definecolor{adminboxborder}{HTML}{F5A623}

\usepackage{indentfirst}
\usepackage[utf8]{inputenc}

\usepackage{makecell}

\title{SLM-SQL: An Exploration of Small Language Models \\for Text-to-SQL}

\author{Lei Sheng\footnotemark[1] \\\makecell{  Wuhan University\\ of Technology, China} \\  \texttt{xuanfeng1992@whut.edu.cn} \\
  \And
  Shuai-Shuai Xu \\ \makecell{ University of Science and \\Technology of China, China} \\ \texttt{sa517432@mail.ustc.edu.cn} \\
}
\usepackage{hyperref}

\begin{document}
\maketitle

\footnotetext[1]{Corresponding author: \texttt{xuanfeng1992@whut.edu.cn}}

\begin{abstract}

Large language models (LLMs) have demonstrated strong performance in translating natural language questions into SQL queries (Text-to-SQL). In contrast, small language models (SLMs) ranging from 0.5B to 1.5B parameters currently underperform on Text-to-SQL tasks due to their limited logical reasoning capabilities. However, SLMs offer inherent advantages in inference speed and suitability for edge deployment. To explore their potential in Text-to-SQL applications, we leverage recent advancements in post-training techniques. Specifically, we used the open-source SynSQL-2.5M dataset to construct two derived datasets: SynSQL-Think-916K for SQL generation and SynSQL-Merge-Think-310K for SQL merge revision. We then applied supervised fine-tuning and reinforcement learning-based post-training to the SLM, followed by inference using a corrective self-consistency approach. Experimental results validate the effectiveness and generalizability of our method, SLM-SQL. On the BIRD development set, the five evaluated models achieved an average improvement of 31.4 points. Notably, the 0.5B model reached 56.87\% execution accuracy (EX), while the 1.5B model achieved 67.08\% EX. We will release our dataset, model, and code to github: \url{https://github.com/CycloneBoy/slm_sql}.
\end{abstract}

\section{Introduction}
\label{sec:introduction}

Converting natural language into SQL for database querying (Text-to-SQL) holds significant application potential and has garnered increasing attention in recent years \citep{katsogiannis-meimarakis_survey_2023, shi_survey_2024,liu_survey_2024}. Large language models (LLMs) exhibit strong capabilities in language understanding and logical reasoning \citep{zhao2025surveylargelanguagemodels}. As a result, most state-of-the-art Text-to-SQL approaches are based on LLMs.

\begin{figure}[t]
	\begin{center}
    \includegraphics[width=.5\textwidth]{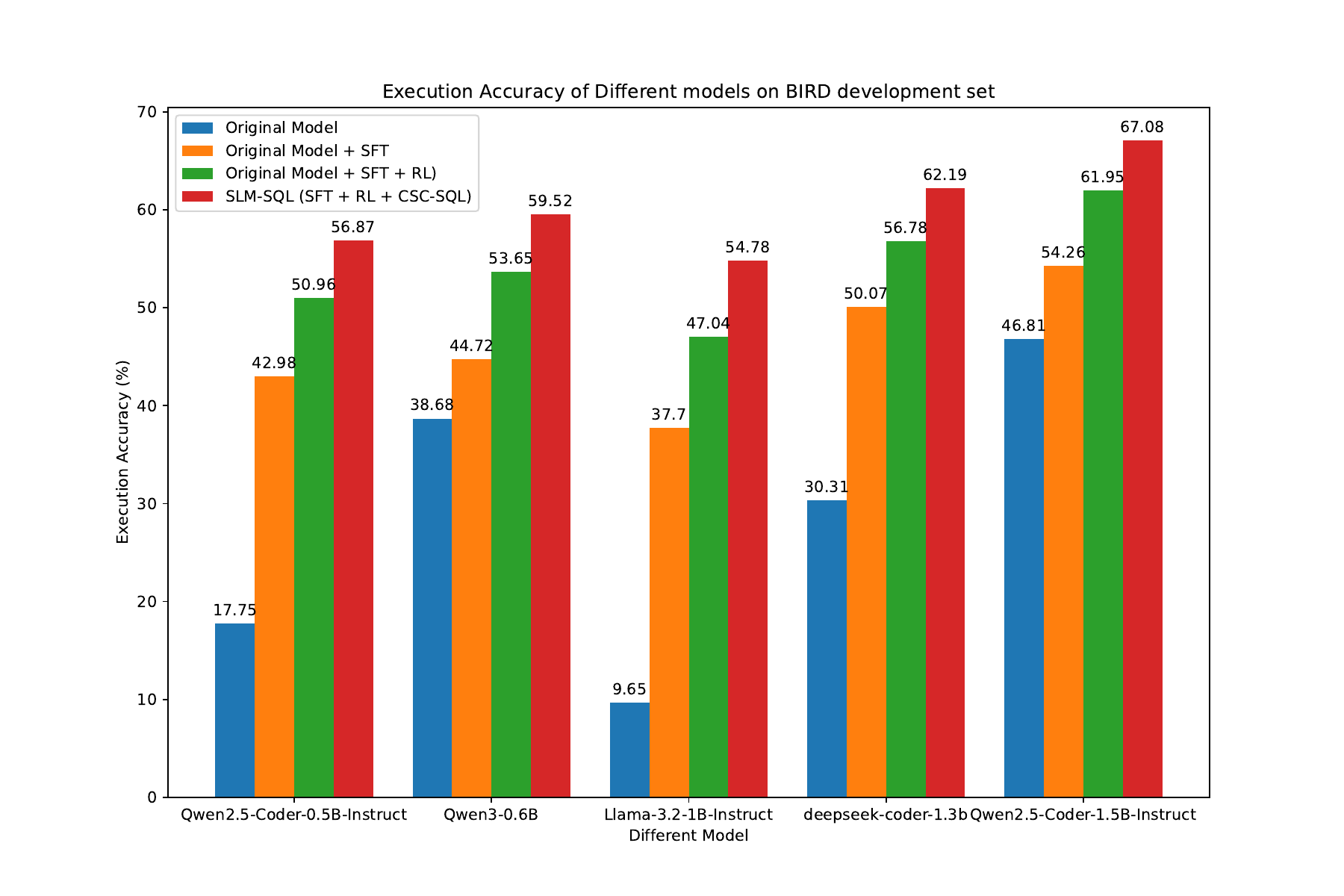}
		  \caption{The execution accuracy of different models on the BIRD development set. The model represented by deepseek-coder-1.3b is deepseek-coder-1.3b-instruct.}
		  \label{figure:slm_sql_with_diff_model_on_bird_dev} 
	\end{center}
\end{figure}

Currently, most open-source LLM-based Text-to-SQL methods utilize models ranging from 3B to 32B parameters, which can be effectively post-trained to enhance their understanding and reasoning capabilities \citep{tie2025surveyposttraininglargelanguage}. Some approaches adopt multi-step pipeline frameworks \citep{pourreza-rafiei-2024-dts,gorti2024msc,sheng2025basesqlpowerfulopensource} , with models primarily fine-tuned using the LoRA method \citep{hu2021lora}. Others leverage pre-training on carefully curated SQL-centric corpora to improve SQL generation capabilities \citep{li_codes_2024,li2025omnisqlsynthesizinghighqualitytexttosql}. Recently, following the success of models such as OpenAI's o1 \citep{openai2024openaio1card} and DeepSeek-R1 \citep{deepseekai2025deepseekr1incentivizingreasoningcapability}, an increasing number of methods \citep{pourreza2025reasoningsqlreinforcementlearningsql,ma2025sqlr1trainingnaturallanguage,papicchio2025think2sqlreinforcellmreasoning,sheng2025cscsqlcorrectiveselfconsistencytexttosql,yao2025arctictext2sqlr1simplerewardsstrong} have adopted reinforcement learning (RL) techniques to further enhance the reasoning abilities of language models.

Recent studies have demonstrated that small language models (SLMs) \citep{vannguyen2024surveysmalllanguagemodels,lu2025smalllanguagemodelssurvey} can also achieve competitive reasoning performance \citep{srivastava2025reasoningabilitysmalllanguage}. However, the application of SLMs to Text-to-SQL tasks remains underexplored. This paper investigates the capabilities of SLMs with 0.5B to 1.5B parameters in Text-to-SQL scenarios. The two-stage generation approach proposed by CSC-SQL \citep{sheng2025cscsqlcorrectiveselfconsistencytexttosql} is relatively simple yet effective; its 3B model achieved 65.28\% execution accuracy (EX) on the BIRD \citep{li_can_2023} development set. Building on this, we propose SLM-SQL, an improved method tailored for smaller models. First, the open-source SynSQL-2.5M dataset \citep{li2025omnisqlsynthesizinghighqualitytexttosql} was processed using heuristic rules to create the SynSQL-Think-916K dataset. Then, inspired by the Corrective Self-Consistency (CSC) technique introduced in CSC-SQL, SynSQL-Think-916K was used to synthesize the SynSQL-Merge-Think-310K dataset. Subsequently, the SQL generation model and SQL merge revision model were trained using supervised fine-tuning (SFT) and reinforcement learning (RL), respectively. Finally, the two-stage inference procedure from CSC-SQL was employed to generate the final SQL outputs.

To evaluate the effectiveness of the SLM-SQL method, we selected five open-source models ranging from 0.5B to 1.5B parameters and conducted experiments on the BIRD and Spider datasets \citep{yu_spider_2019}. The experimental results are presented in Figure \ref{figure:slm_sql_with_diff_model_on_bird_dev}. On the BIRD development set, the five models achieved an average improvement of 31.4 points, with the 0.5B model reaching 56.87\% execution accuracy (EX) and the 1.5B model achieving 67.08\% EX, demonstrating the effectiveness of SLM-SQL. We then directly tested the models trained on the BIRD dataset on the Spider dataset, where all five models also exhibited significant performance gains. The 0.5B model achieved 73.50\% EX, while the 1.5B model reached 79.06\% EX, indicating that SLM-SQL generalizes well across datasets.

Our contributions are as follows:

(1) Based on the SynSQL-2.5M dataset, we constructed two datasets: SynSQL-Think-916K for SQL generation and SynSQL-Merge-Think-310K for SQL merge revision.

(2) We applied supervised fine-tuning and reinforcement learning to enhance the SQL generation capabilities of SLMs and trained five different models to evaluate the effectiveness and generalization of the proposed SLM-SQL method.

(3) Our method achieved 56.87\% EX for the 0.5B model and 67.08\% EX for the 1.5B model on the BIRD development set, demonstrating competitive performance.

\section{Methodology}
\label{sec:methods}

\begin{figure}[htbp]
	\begin{center}
		  \includegraphics[width=0.5\textwidth]{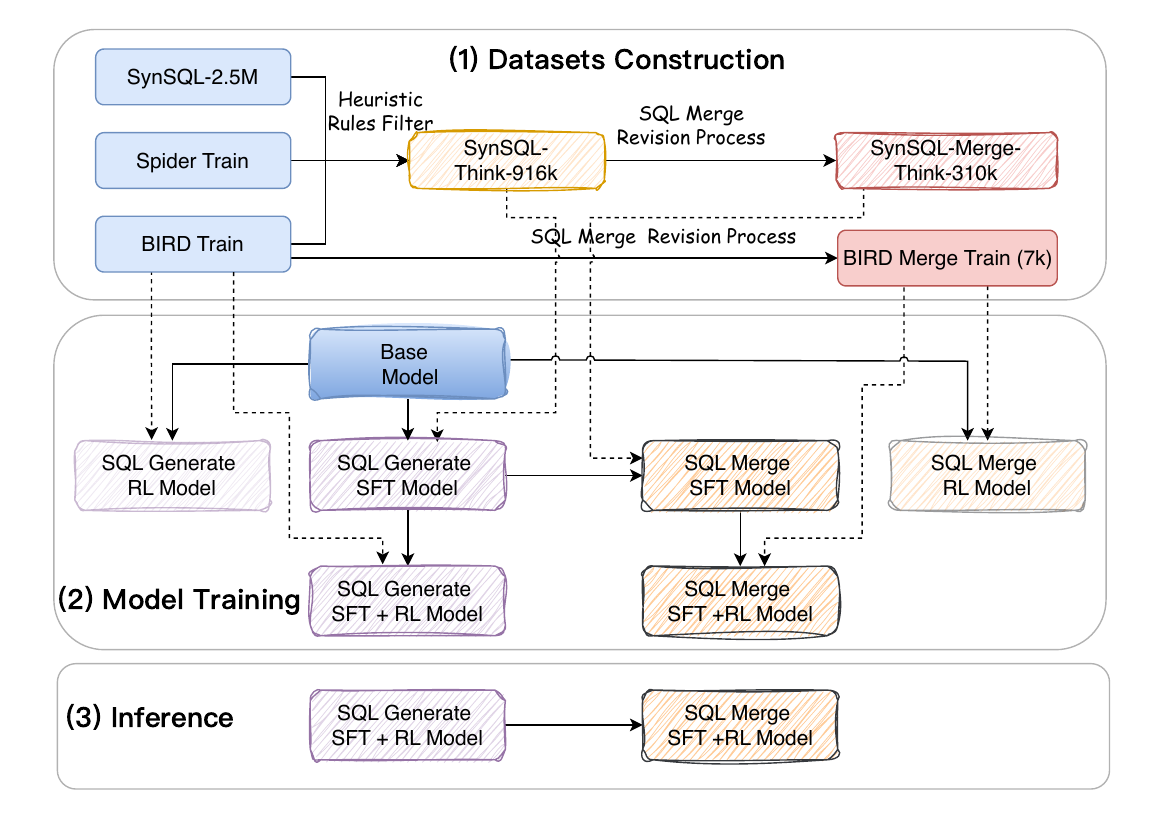}
		  \caption{Overview of the proposed SLM-SQL framework.}
		  \label{figure:slm_sql} 
	\end{center}
\end{figure}

Our SLM-SQL framework is illustrated in Figure \ref{figure:slm_sql} and comprises three main components: dataset construction, model training, and inference.

\textbf{Dataset Construction:}: We use the SynSQL-2.5M dataset \citep{li2025omnisqlsynthesizinghighqualitytexttosql} as the primary data source. This large-scale, high-quality, and diverse synthetic dataset was specifically developed for the Text-to-SQL task. It also includes the Spider and BIRD training sets, for which the OmniSQL method \citep{li2025omnisqlsynthesizinghighqualitytexttosql} has generated Chain-of-Thought (CoT) annotations. To reduce the learning difficulty and dataset size, we apply heuristic preprocessing rules.
First, we filter out samples that (i) do not contain the SELECT keyword, (ii) the SQL statement appeared multiple times in the CoT, or (iii) include the '--' comment in the SQL statement.
Next, we clean the CoT by removing content after the SQL statement (typically explanations or reflections) and place the preceding portion between <think> and </think> tags. We also remove markdown-specific SQL tags and enclose the SQL statements between <answer> and </answer> tags.
Finally, we discard samples with input prompt lengths is higher than 7,000 tokens, resulting in the SynSQL-Think-916K dataset.

We employed the CSC-SQL method to construct a new merge revision dataset, SynSQL-Merge-Think-310K. First, we used the Qwen2.5-Coder-7B-Instruct model to generate eight candidate outputs in parallel on the SynSQL-Think-916K dataset. We then performed group voting based on SQL execution results and selected the two groups with the highest vote counts to create the merge revision dataset.

\textbf{Model Training:} In the CSC-SQL method, both the SQL generation and SQL merge revision models are directly post-trained using RL. In contrast, our SLM-SQL method draws on SQL-R1 \citep{ma2025sqlr1trainingnaturallanguage} and Think2SQL \citep{papicchio2025think2sqlreinforcellmreasoning}, applying SFT on synthetic data before RL-based post-training to enhance the SQL generation capabilities of SLMs.

First, we select a SLM with 0.5B to 1.5B parameters and perform SFT on the SQL generation dataset SynSQL-Think-916K. The training objective is to minimize the cross-entropy loss between the predicted CoT and the synthesized CoT, enabling the model to first generate the reasoning process and then generate SQL. Next, we perform RL-based post-training on the BIRD training set using the Group Relative Policy Optimization (GRPO) \citep{shao2024deepseekmathpushinglimitsmathematical} algorithm, further improving the SQL generation performance. The reward function consists of two components: execution accuracy reward ($R_{EX}$) and format reward ($R_{Format}$). See Appendix \ref{sec:reward_design} for details.

For the merge revision model, we first apply SFT on the SynSQL-Merge-Think-310K dataset using the SFT-trained SQL generation model. We then conduct GRPO-based post-training on the BIRD-Merge-Train dataset, which is synthesized using the CSC-SQL method.

\textbf{Inference:} The inference procedure of our SLM-SQL method follows the same approach as the CSC-SQL method. First, the SQL generation model performs parallel sampling. Then, the generated SQL statements are grouped based on their execution results through a voting mechanism. If the voting results are inconsistent, the SQL is regenerated using the SQL merge revision model; otherwise, the SQL with consistent votes is selected as the final output.

\begin{table}[htbp]\tiny
  \setlength{\tabcolsep}{4pt}
  \centering
\begin{tabular}{c|c|c|c}    
  \toprule
  \text{Dataset} & \text{Train Model} & \makecell[c]{Train \\Method} & \text{Size} \\
  \midrule 
  \makecell[l]{SynSQL-2.5M \citep{li2025omnisqlsynthesizinghighqualitytexttosql}}  & \multirow{3}*{\makecell[c]{SQL Generate \\Model}} & \text{SFT} & \text{2,190,988}\\
  \makecell[l]{SynsQL-Think-916k}  &  & \text{SFT} & \text{916,156}\\
  \makecell[l]{BIRD Train \citep{li2024can}}  &  & \text{RL} & \text{9428}\\
  \midrule
  \makecell[l]{SynsQL-Merge-Think-310k}  & \multirow{2}*{\makecell[c]{SQL Merge \\Revision Model}} & \text{SFT} & \text{310,764}\\
  \makecell[l]{BIRD Merge Train \citep{sheng2025cscsqlcorrectiveselfconsistencytexttosql}}  &  & \text{RL} & \text{7159}\\

  \bottomrule
\end{tabular}
\caption{Statistics of different datasets.}
\label{tab:dataset_slm_sql}
\footnotetext{}
\end{table}

\section{Experiments}
\label{sec:experiments}

\subsection{Experiments Setting}
\label{subsec:experiments_setting}

We conduct experiments on the BIRD and Spider datasets, using the widely adopted execution accuracy (EX) metric to evaluate the performance of our framework. The statistics of all experimental datasets are presented in Table \ref{tab:dataset_slm_sql}. Five SLMs ranging in size from 0.5B to 1.5B were selected as base models for the experiments. Additional implementation details can be found in Appendix \ref{sec:appendix_implementation_details}.

\begin{table}[htbp]\tiny
  \setlength{\tabcolsep}{4pt}
  \centering
\begin{tabular}{c|c|c|c|c}    
  \toprule
  \text{Method} & \text{Model} & \text{Size} & \makecell[c]{Dev \\EX(\%)} & \makecell[c]{Test \\EX(\%)} \\
  \midrule 
  \makecell[l]{AskData\\ \citep{shkapenyuk2025automaticmetadataextractiontexttosql}} & \text{GPT-4o} & \text{UNK} &\text{75.36} &\text{77.14}\\
  \makecell[l]{CHASE-SQL \\ \citep{pourreza2024chasesqlmultipathreasoningpreference}} & \text{Gemini-1.5-pro} & \text{UNK} &\text{73.01} & \text{73.0}\\
  \makecell[l]{RSL-SQL\\ \citep{cao2024rslsqlrobustschemalinking}} & \text{GPT-4o} & \text{UNK} &\text{67.21} &\text{68.70}\\
  \makecell[l]{MCS-SQL\\ \citep{lee_mcs-sql_2024}} & \text{GPT-4} & \text{UNK} &\text{63.36} &\text{65.45}\\

  \midrule 
  
  \makecell[l]{Reasoning-SQL\\ \citep{pourreza2025reasoningsqlreinforcementlearningsql}} & \text{Qwen2.5-Coder-14B-Instruct} & \text{14B} &\text{72.29} &\text{72.78}\\

  \makecell[l]{OMNI-SQL \\ \citep{li2025omnisqlsynthesizinghighqualitytexttosql}} & \text{Qwen2.5-Coder-32B-Instruct} & \text{32B} &\text{69.23} &\text{72.05}\\
  

  \makecell[l]{Alpha-SQL\\ \citep{li2025alphasqlzeroshottexttosqlusing}} & \text{Qwen2.5-Coder-32B-Instruct} & \text{32B} &\text{69.70} &\text{70.26}\\

  \makecell[l]{XiYan-SQL\\ \citep{xiyansql}} & \makecell[c]{XiYanSQL-QwenCoder\\-32B-2412} & \text{32B} &\text{67.01} &\text{69.03}\\

  \midrule 

  \makecell[l]{Arctic-Text2SQL-R1\\ \citep{yao2025arctictext2sqlr1simplerewardsstrong}} & \text{OmniSQL-7B} & \text{7B} &\text{68.90} &\text{68.47}\\




  \makecell[l]{SQL-R1 \\ \citep{ma2025sqlr1trainingnaturallanguage}} & \text{Qwen2.5-Coder-7B-Instruct} & \text{7B} &\text{66.60} &\text{-}\\

  \makecell[l]{OMNI-SQL \\ \citep{li2025omnisqlsynthesizinghighqualitytexttosql}} & \text{Qwen2.5-Coder-7B-Instruct} & \text{7B} &\text{66.10} &\text{67.97}\\

  \makecell[l]{CSC-SQL\\ \citep{sheng2025cscsqlcorrectiveselfconsistencytexttosql}} & \makecell[c]{XiYanSQL-QwenCoder\\-3B-2502} & \text{3B} &\text{65.28} &\text{-} \\

  \makecell[l]{CodeS \\ \citep{li_codes_2024}} & \text{StarCoder} & \text{15B} &\text{58.47} &\text{60.37}\\

  \makecell[l]{DTS-SQL \\ \citep{pourreza-rafiei-2024-dts}} & \text{DeepSeek 7B} & \text{7B} &\text{55.80} &\text{60.31}\\

  \makecell[l]{Prem-1B-SQL \\ \citep{Anindyadeep2024PremSQL}} & \text{deepseek-coder-1.3b-instruct} & \text{1.3B} &\text{46.0} &\text{51.54}\\

  \makecell[l]{Qwen2.5-Coder\\ \citep{hui2024qwen2}} & \text{Qwen2.5-Coder-1.5B-Instruct} & \text{1.5B} &\text{28.40} &\text{-} \\
  
  \midrule 

  \multirow{5}*{\makecell[l]{SLM-SQL (Our)}} & \text{Qwen2.5-Coder-0.5B-Instruct} & \text{0.5B} &\text{56.87} &\text{-} \\
   & \text{Qwen3-0.6B} & \text{0.6B} &\text{59.52} &\text{-} \\
 & \text{Llama-3.2-1B-Instruct} & \text{1B} &\text{54.78} &\text{-} \\
   & \text{deepseek-coder-1.3b-instruct} & \text{1.3B} &\text{62.19} &\text{-} \\
   & \text{Qwen2.5-Coder-1.5B-Instruct} & \text{1.5B} &\text{67.08} &\text{-} \\

  \bottomrule
\end{tabular}
\caption{Performance Comparison of different Text-to-SQL methods on BIRD dev  and private test dataset.}
\label{tab:result_main_bird_other}
\footnotetext{}
\end{table}

\subsection{Main Results}
\label{subsec:main_result}

\begin{table}[htbp]\tiny
  \centering
\begin{tabular*}{0.875\linewidth}{c|c|c|c|c|c}    
  \toprule
 \multirow{2}*{\text{Model}}   & \multirow{2}*{\makecell{Train \\Method}}  &  \multicolumn{2}{c|}{\text{Dev EX(\%)}} &  \multicolumn{2}{c}{\text{Test EX(\%)}} \\
 \cline{3-6}
 & & \text{SC} & \text{CSC} & \text{SC} & \text{CSC}\\
\midrule 
  \multirow{3}*{\makecell[l]{Qwen2.5-Coder\\-0.5B-Instruct}}  & \text{-} & \text{42.13} & \text{44.07} & \text{42.23} & \text{44.31} \\
& \text{SFT} & \text{65.31} & \text{68.28} & \text{67.26} & \text{70.72} \\
&  \text{SFT + RL} & \text{70.60} & \text{72.08} & \text{70.42} & \text{73.50} \\
\midrule

\multirow{3}*{\makecell[l]{Qwen3-0.6B}}   & \text{-} & \text{63.19} & \text{65.15} & \text{64.20} & \text{66.03} \\
& \text{SFT} & \text{68.02} & \text{70.05} & \text{71.48} & \text{73.90} \\
& \text{SFT + RL} & \text{72.05} & \text{72.99} & \text{73.89} & \text{75.81} \\
\midrule
\multirow{3}*{\makecell[l]{Llama-3.2\\-1B-Instruct }}   & \text{-} & \text{30.79} & \text{32.72} & \text{32.71} & \text{34.73} \\
& \text{SFT} & \text{63.77} & \text{66.18} & \text{69.06} & \text{70.86} \\
& \text{SFT + RL} & \text{67.63} & \text{69.12} & \text{73.05} & \text{74.29} \\
\midrule
\multirow{3}*{\makecell[l]{deepseek-coder\\-1.3b-instruct }}  & \text{-} & \text{48.23} & \text{49.94} & \text{49.09} & \text{51.19} \\
& \text{SFT} & \text{72.86} & \text{74.31} & \text{75.13} & \text{77.19} \\
& \text{SFT + RL} & \text{75.47} & \text{76.11} & \text{77.05} & \text{78.08} \\
\midrule
\multirow{3}*{\makecell[l]{Qwen2.5-Coder\\-1.5B-Instruct}}  & \text{-} & \text{63.54} & \text{65.99} & \text{67.61} & \text{69.34} \\
& \text{SFT} & \text{74.53} & \text{76.66} & \text{77.35} & \text{79.13} \\
& \text{SFT + RL} & \text{75.15} & \text{76.72} & \text{78.42} & \text{79.06} \\
\bottomrule
\end{tabular*}
\caption{The table shows the EX comparison results of different models on the Spider dataset. SC stands for direct use of Self-Consistency, and CSC stands for Corrective Self-Consistency. The number of SQL generation model samples is 16, and the number of SQL merge revision model samples is 8. The merge revision model of the CSC method uniformly uses the Qwen2.5-Coder-0.5B-Instruct model trained after SFT and RL.}
\label{tab:result_main_spider}
\footnotetext{}
\end{table}

\textbf{BIRD Results:}  Table \ref{tab:result_main_bird_other} presents the evaluation results of SLM-SQL and baseline methods on the BIRD dataset. On the BIRD development set, SLM-SQL achieved competitive performance across model sizes ranging from 0.5B to 1.5B. Notably, the 0.6B model attained 59.52\% EX, outperforming the 7B DTS-SQL \citep{pourreza-rafiei-2024-dts} and the 15B CodeS \citep{li_codes_2024} models. The 1.5B variant of SLM-SQL achieved 67.08\% EX, surpassing several significantly larger models, including Qwen2.5-Coder-7B-Instruct and XiYanSQL-QwenCoder-32B. It also outperformed closed-source LLM-based methods such as RSL-SQL \citep{cao2024rslsqlrobustschemalinking} and MCS-SQL \citep{lee_mcs-sql_2024}, which rely on GPT-4 and GPT-4o, respectively. Compared to CSC-SQL (3B), SLM-SQL-1.5B improves performance by 1.8 points. Additionally, compared to Prem-1B-SQL \citep{Anindyadeep2024PremSQL} which is based on the same DeepSeek-Coder-1.3B-Instruct model SLM-SQL-1.3B shows an improvement of 16.19 points, further demonstrating the effectiveness of our approach.

\textbf{Spider Results:} To evaluate the generalization capability of the SLM-SQL method, we directly tested models trained on the BIRD dataset using the Spider dataset. The results are presented in Table \ref{tab:result_main_spider}. Five SLM-SQL models of varying sizes achieved competitive performance on the Spider dataset, including an average improvement of 23.83 percentage points on the development set and 24.98 percentage points on the test set. Specifically, SLM-SQL-0.5B achieved 72.08\% EX on the development set and 73.5\% EX on the test set, while SLM-SQL-1.5B achieved 76.72\% EX and 79.06\% EX, respectively. Notably, despite not undergoing RL post-training on the Spider dataset, SLM-SQL still achieved strong performance, demonstrating robust generalization.

\subsection{Ablation Study}
\label{sec:ablation_study}

\begin{table}[htbp]\tiny
  \centering
\begin{tabular*}{0.875\linewidth}{l|c|c|c|c}    
  \toprule
 \multirow{2}*{\text{Method}} & \multicolumn{4}{c}{Dev EX(\%)} \\
 \cline{2-5}
 & \text{0.5B} & ${\Delta}$\text{EX} & \text{1.5B} & ${\Delta}$\text{EX}\\
\midrule
\text{SLM-SQL} & 56.87 & - & 67.08 & - \\
\hline
\makecell[l]{with SQL generation \\ model samples is 16} & 53.22 & \text{-3.64} & 64.84 & -2.24 \\
\hline
\text{w/o Corrective Self-Consistency} & 50.96 & -5.91 & 61.95 & -5.13 \\
\text{w/o RL} & 51.52 & -5.35 & 62.91 & -4.17 \\
\text{w/o SFT} & 34.94 & -21.93 & 58.19 & -8.89 \\
\text{w/o SFT and RL} & 22.14 & -34.73 & 52.26 & -14.82 \\

\bottomrule
\end{tabular*}
\caption{Ablation study of SLM-SQL on BIRD development set. 0.5B and 1.5B represent the basic models used, namely Qwen2.5-Coder-0.5B-Instruct and Qwen2.5-Coder-1.5B-Instruct.}
\label{tab:result_ablation_bird_dev}
\footnotetext{}
\end{table}

Table \ref{tab:result_ablation_bird_dev} presents the results of our ablation study. Removing any module results in a significant performance degradation. In particular, excluding the SFT module causes the performance of SLM-SQL-0.5B to drop by 21.93 points and that of SLM-SQL-1.5B by 8.89 points, highlighting the importance of synthetic data for SFT. Removing the Corrective Self-Consistency module leads to an approximate 5-point drop in performance, suggesting that the merge revision model trained on SLMs remains effective in error correction. Additionally, reducing the number of SQL samples from 64 to 16 results in a performance drop of 3.64 points for SLM-SQL-0.5B and 2.24 points for SLM-SQL-1.5B, demonstrating the benefit of increased computational budgets during inference.

\section{Conclusion}
\label{sec:conclusion}

In this study, we investigate the performance of several 0.5B-1.5B SLMs on Text-to-SQL tasks. To address the limited understanding and reasoning capabilities of small language models, we reorganize the SynSQL-Think-916k dataset for SQL generation and the SynSQL-Merge-Think-310k dataset for SQL merge revision. We then enhance the reasoning ability of SLM-SQL through a combination of supervised fine-tuning and reinforcement learning. Experimental results demonstrate that SLMs in the 0.5B-1.5B range hold significant promise for solving Text-to-SQL tasks.

\section{Limitations}
\label{sec:limitations_and_future_work}

In this paper, we focus exclusively on evaluating the performance of SLMs in the Text-to-SQL task using Self-Consistency and Corrective Self-Consistency methods, without exploring other more advanced approaches such as schema linking, agent-based frameworks, or pipeline-based methods. 
We plan to investigate these areas in future work. Moreover, our current study is limited to the Text-to-SQL domain, and extending this research to broader code generation tasks represents an important future direction.

\section{Ethical considerations}
\label{sec:ethical_statement}

All datasets and models used in this study are publicly available. We will release the processed datasets, trained models, and source code after the review process to promote transparency and reproducibility. Additionally, our work focuses solely on Text-to-SQL generation, which does not involve the production of harmful or biased content. The synthetic dataset was generated using a large language model without human annotation, thereby ensuring that no forced labor was involved.

\newpage
\bibliography{custom}

\appendix

\newpage


\section{Related Work}
\label{sec:related_works}

\begin{table*}[htbp]\tiny
  \setlength{\tabcolsep}{4pt}
  \centering
\begin{tabular}{c|c|c|c|c|c|c|c|c|c|c}    
  \toprule
  \text{Model} & \text{Model Size} & \makecell{Train\\Method} & \text{SC Size} & \makecell{SC Use \\time (Hour)} & \makecell{CSC Use \\Time (Hour)} & \makecell{The proportion \\ of time that \\CSC increases (\%)} & \makecell{Total Use \\ Time (Hour)} & \makecell{Total \\Cost (\$) } & \makecell{Average Cost\\ Per Question (\$)} & \makecell[c]{Bird \\Dev EX (\%)}  \\
  \toprule
  \multirow{2}*{\text{\makecell{SLM-SQL-1.5B}}} & \multirow{2}*{\text{1.5B}} & \multirow{2}*{\text{SFT + RL + CSC}}
      & 16 & 0.63 & 0.2 & 32 & 0.83  & 0.22 & 0.00014 & 64.84 \\
  & & & 64 & 2.43 & 0.25 & 10 & 2.68 & 0.7  & 0.00046 & 67.08 \\
  \bottomrule
\end{tabular}
\caption{The table shows the cost and performance of SQL generation on the BIRD development set using our SLM-SQL-1.5B. The inference cost analysis uses an NVIDIA 4090D with 24GB of memory (rented at \$0.26 per hour) as an example.}
\label{tab:result_cost_and_performance}
\footnotetext{}
\end{table*}

\textbf{Text-to-SQL } methods have evolved from early rule-based approaches and fine-tuned pre-trained language models \citep{wang-etal-2020-rat,guo-etal-2019-towards} to large language model (LLM)-based approaches \citep{liu_survey_2024,hong_next-generation_2024}. These approaches can be categorized into three groups: (1) in-context learning (ICL)-based methods \citep{dong_c3_2023,pourreza_din-sql_2023,gao_text--sql_2023,lee_mcs-sql_2024}, (2) methods based on fine-tuning open-source LLMs \citep{pourreza-rafiei-2024-dts,li_codes_2024,sheng2025basesqlpowerfulopensource,li2025omnisqlsynthesizinghighqualitytexttosql}, and (3) hybrid approaches combining ICL and supervised fine-tuning (SFT) \citep{talaei_chess_2024,pourreza2024chasesqlmultipathreasoningpreference, xiyansql}. With the emergence of Test-Time Scaling techniques \citep{zhang2025surveytesttimescalinglarge}, several works have incorporated strategies such as self-consistency (SC) \citep{gao_text--sql_2023,xie2025opensearchsqlenhancingtexttosqldynamic,sheng2025cscsqlcorrectiveselfconsistencytexttosql}, self-correction \citep{pourreza2024chasesqlmultipathreasoningpreference,xiyansql}, and Monte Carlo Tree Search (MCTS) \citep{yuan2025mctssqleffectiveframeworktexttosql,lyu2025sqlo1selfrewardheuristicdynamic,li2025alphasqlzeroshottexttosqlusing} to enhance generation performance. 
Recently, reinforcement learning (RL) for post-training has proven effective in improving the reasoning capabilities of LLMs  \citep{openai2024openaio1card,deepseekai2025deepseekr1incentivizingreasoningcapability,yang2025qwen3technicalreport}, and an increasing number of methods \citep{pourreza2025reasoningsqlreinforcementlearningsql,sheng2025cscsqlcorrectiveselfconsistencytexttosql} have adopted this technique. Notably, \citep{ma2025sqlr1trainingnaturallanguage,papicchio2025think2sqlreinforcellmreasoning,yao2025arctictext2sqlr1simplerewardsstrong} first leverage synthetic datasets for SFT, followed by RL-based post-training, which significantly enhances SQL generation performance.

\textbf{Small language models} (SLMs) have gained increasing attention due to their efficiency and strong performance \citep{vannguyen2024surveysmalllanguagemodels}. \citep{srivastava2025reasoningabilitysmalllanguage} conducted a comprehensive analysis of the reasoning capabilities of various SLMs, while \citep{anjum2025reasoningbeatsscale15b} evaluated the performance of the distilled reasoning model DeepSeek-R1-1.5B on the Text-to-SQL task. In addition, CSC-SQL \citep{sheng2025cscsqlcorrectiveselfconsistencytexttosql} introduced a merge revision module and ultimately employed a 3B model to achieve an execution accuracy (EX) of 65.28\% on the BIRD development set.

\begin{figure}[htbp]
	\begin{center}
    \includegraphics[width=.5\textwidth]{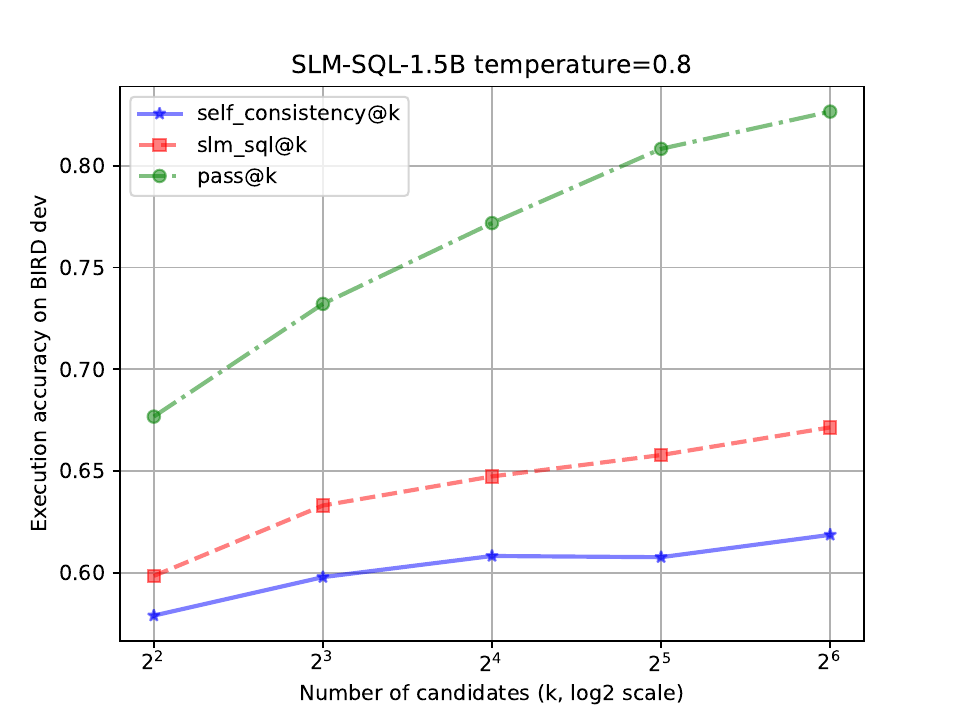}
		  \caption{Trend chart of various metrics of SLM-SQL-1.5B under different sampling numbers on the BIRD development set. self\_consistency@k and slm\_sql@k represent the results of using the self-consistency method and SLM-SQL method respectively.}
		  \label{figure:slm_sql_1.5b_temperature_0.8} 
	\end{center}
\end{figure}

\section{Implementation details}
\label{sec:appendix_implementation_details}

We employed five models as the foundation for our experiments: Qwen2.5-Coder-0.5B-Instruct, Qwen3-0.6B \citep{yang2025qwen3technicalreport}, Llama-3.2-1B-Instruct \citep{grattafiori2024llama3herdmodels}, DeepSeek-Coder-1.3B-Instruct \citep{deepseek-coder}, and Qwen2.5-Coder-1.5B-Instruct. Based on these models, we trained corresponding SQL generation models. Additionally, we trained two merge revision models using Qwen2.5-Coder-0.5B-Instruct and Qwen2.5-Coder-1.5B-Instruct.

For supervised fine-tuning (SFT) of SLM-SQL, the training configuration included a learning rate of 2.0e-5 with linear scheduling, a warm-up rate of 0.1, an effective batch size of 1024, and training over 2 epochs. For reinforcement learning (RL), we adopted the GRPO algorithm with a learning rate of 3e-6, a cosine learning rate scheduler, a warm-up rate of 0.1, and 1 training epoch. GRPO rollouts were set to 6, with clip ratios of 0.2 and 0.28. All experiments were conducted on a machine equipped with four NVIDIA GPUs, each with 80 GB of VRAM. For additional implementation details, please refer to the open-source code.

Unless otherwise specified, each SQL generation experiment produces 64 samples, while the merge revision process generates 8 samples. We report the average performance over three independent runs for each experimental setting. During inference, SLM-SQL employs a 0.5B-sized merge revision model for all three base models: Qwen3-0.6B, Llama-3.2-1B-Instruct, and DeepSeek-Coder-1.3B-Instruct.

\section{Reward Design}
\label{sec:reward_design}

During GRPO post-training for the Text-to-SQL task, we adopt a simple reward function composed of two components: execution accuracy reward ($R_{\text{EX}}$) and format reward ($R_{\text{Format}}$).

\begin{align}
R_{EX}= 
\begin{cases} 
1, & \text{if execution results is correct.} \\
0, & \text{otherwise}
\end{cases}
\end{align}

\begin{align}
R_{Format}= 
\begin{cases} 
1, & \text{if output format is match.} \\
0, & \text{otherwise}
\end{cases}
\end{align}

The final reward is calculated as the weighted sum of $R_{\text{EX}}$ and $R_{\text{Format}}$ :

\begin{equation}
  R = R_{EX} + 0.1 * R_{Format}
\end{equation}

\section{Additional Analysis}
\label{sec:additional_analysis}

\subsection{Test-time Compute Analysis}
\label{subsec:analysis}

We further analyzed the impact of varying the number of samples on the performance of SLM-SQL. Experiments were conducted on the BIRD development set with sample sizes ranging from 4 to 64, as illustrated in Figure \ref{figure:slm_sql_1.5b_temperature_0.8}. The results indicate that increasing the number of samples improves the EX of SLM-SQL-1.5B. While the performance gains from the self-consistency method plateau beyond 16 samples, the SLM-SQL method continues to show significant improvement, suggesting its superior effectiveness. The pass@k metric also increases notably with larger sample sizes, reaching up to 0.8264 for pass@64. These findings demonstrate the strong potential of SLMs in solving the Text-to-SQL task.

Furthermore, increasing the number of generated samples leads to higher inference time and computational cost. Table \ref{tab:result_cost_and_performance} presents the inference time and cost of SLM-SQL-1.5B. Using an NVIDIA 4090D GPU (rental cost approximately \$0.26 per hour), increasing the number of generated samples from 16 to 64 raises the total inference time on the BIRD development set from 0.83 to 2.68 hours, and the average cost per question from \$0.00014 to \$0.00046. This results in roughly a threefold increase in both time and cost, while the EX improves from 64.84\% to 67.08\%. In practical applications, setting the number of samples to 16 offers a balanced trade-off between performance and efficiency.

\subsection{Difficulty Analysis}
\label{sec:difficulty_analysis}

Table \ref{tab:result_main_sep_bird} presents the performance of SLM-SQL-0.5B and SLM-SQL-1.5B across varying difficulty levels using different training methods on the BIRD development set. The results demonstrate that each training method (SFT, RL, and CSC) yields notable performance improvements depending on the difficulty level.

\onecolumn
\begin{table*}[htbp]\small
  \centering
\begin{tabular*}{0.7\linewidth}{c|c|l|l|l|l}    
  \toprule

  \multirow{2}*{\text{Model}}  & \multirow{2}*{\makecell{Train\\Method}}   &  \multicolumn{4}{c}{\text{Dev EX(\%)}} \\
  \cline{3-6}
  & & \text{Simple} & \text{Moderate} & \text{Challenge} & \text{All} \\ 
  \midrule

  \multirow{4}*{\text{\makecell{SLM-SQL-0.5B}}}  & \text{-}  & \text{25.41} & \text{7.11} & \text{4.14} & \text{17.86}  \\
  & \text{SFT}  & \text{53.30} & \text{30.60 } & \text{17.93} & \text{43.09}  \\
 & \text{SFT + RL}  & \text{60.54} & \text{38.36} & \text{26.21} & \text{50.59}  \\
 & \text{SFT+ RL+ CSC} & \text{65.51} & \text{46.77} & \text{31.72} & \text{56.65}  \\
  \midrule
  \multirow{4}*{\text{\makecell{SLM-SQL-1.5B}}}  & \text{-}  & \text{56.43} & \text{33.41} & \text{28.28} & \text{46.81}  \\
  & \text{SFT}  & \text{61.73} & \text{43.97} & \text{38.62} & \text{54.17}  \\
 & \text{SFT + RL}  & \text{69.19} & \text{52.37} & \text{45.52} & \text{50.59}  \\
 & \text{SFT+ RL+ CSC} & \text{73.30} & \text{59.91} & \text{51.03} & \text{67.14}  \\

  \bottomrule
\end{tabular*}
\caption{The performance of SLM-SQL at different difficulty levels on the BIRD development set.}
\label{tab:result_main_sep_bird}
\footnotetext{}
\end{table*}

\newpage
\section{Output Example}
\label{sec:output_example_analysis}

In this section, we show examples of the output results of the Qwen2.5-Coder-1.5B-Instruct model using different training methods.


\begin{tcolorbox}[colback=mainboxbg, colframe=mainboxborder, coltitle=black, fonttitle=\bfseries, title=Qwen2.5-Coder-1.5B-Instruct
 original model output results for question 12 ,bottom=50mm, boxrule=0.5mm, arc=2mm, outer arc=2mm]
<think>\\
To solve this question, we need to follow these steps:\\

1. **Identify the relevant tables**: We need to access data from the `frpm` table for the SAT scores and the `satscores` table for the eligibility rates.\\

2. **Determine the join condition**: Since the `frpm` table and the `satscores` table share the common column `cds`, we will join these two tables on this column.\\

3. **Filter the data for schools with an SAT excellence rate over 0.3**: We will use a WHERE clause to filter the rows where the `Excellence rate` is greater than 0.3.\\

4. **Calculate the eligible free rate for students aged 5-17**: We will use the formula `Eligible free rate = `Free Meal Count (Ages 5-17)` / `Enrollment (Ages 5-17)`. We will create a subquery to calculate this for each school and then join this subquery with the main `frpm` table to get the corresponding district name.\\

5. **Order the results by the eligible free rate in descending order**: To get the highest eligible free rate, we will order the results by the `Eligible free rate` column in descending order.\\

6. **Limit the results to the top school**: Since we are only interested in the highest eligible free rate, we will limit the results to the top 1 row.\\

7. **Select the required columns**: We will select the `School`, `District Name`, `Eligible free rate`, and `Excellence rate` columns from the result set.\\

By following these steps, we can generate the final SQL query.\\
</think>\\
<answer>SELECT T2.School, T2.District Name, CAST(T2.`Free Meal Count (Ages 5-17)` AS REAL) / T2.`Enrollment (Ages 5-17)` AS Eligible\_free\_rate, T1.Excellence\_rate FROM frpm AS T1 INNER JOIN satscores AS T2 ON T1.cds = T2.cds WHERE T1.`Excellence rate` > 0.3 ORDER BY Eligible\_free\_rate DESC LIMIT 1\\
</answer>\\
\end{tcolorbox}


\begin{tcolorbox}[colback=mainboxbg, colframe=mainboxborder, coltitle=black, fonttitle=\bfseries, title=Qwen2.5-Coder-1.5B-Instruct
 after SFT output results for question 223 ,bottom=5mm, boxrule=0.5mm, arc=2mm, outer arc=2mm]
<think> \\
To translate the natural language question into an executable SQLite query, let's break it down step by step:\\


\#\#\# Step 1: Understand the Question\\
The question asks for the **atom IDs** associated with a specific bond, identified by `TR000\_2\_5`. The bond is described in the `bond` table, and the atom IDs are stored in the `connected` table.\\


\#\#\# Step 2: Identify Relevant Tables\\
1. **`bond` Table**:\\
   - This table contains information about bonds, including their `bond\_id`.\\
   - We need to find the `atom\_id1` and `atom\_id2` associated with `TR000\_2\_5`.\\

2. **`connected` Table**:\\
   - This table links `atom\_id1`, `atom\_id2`, and `bond\_id`.\\
   - We will use this table to find the atom IDs that are part of the bond identified by `TR000\_2\_5`.\\


\#\#\# Step 3: Define the Relationship Between Tables\\
- The `bond` table and the `connected` table are related through the `bond\_id` field.\\
- Specifically, `bond.bond\_id` is the foreign key in the `connected` table, linking it to the `bond` table.\\


\#\#\# Step 4: Filter for the Specific Bond\\
- We need to filter the `bond` table for the row where `bond\_id = 'TR000\_2\_5'`.\\


\#\#\# Step 5: Retrieve Atom IDs\\
- After filtering the `bond` table, we need to retrieve the `atom\_id1` and `atom\_id2` associated with this bond.\\
- The `connected` table will help us join the `atom\_id` values to the bond.\\


\#\#\# Step 6: Write the SQL Query\\
We will use a `JOIN` to combine the `bond` and `connected` tables, and then filter for the specific bond `TR000\_2\_5`. Finally, we select the `atom\_id1` and `atom\_id2`.\\

Here is the SQL query:\\
</think>\\
<answer>\\
SELECT connected.atom\_id1, connected.atom\_id2 FROM bond JOIN connected ON bond.bond\_id = connected.bond\_id WHERE bond.bond\_id = 'TR000\_2\_5';\\
</answer>\\
\end{tcolorbox}


\begin{tcolorbox}[colback=mainboxbg, colframe=mainboxborder, coltitle=black, fonttitle=\bfseries, title=Qwen2.5-Coder-1.5B-Instruct
 after SFT and RL output results for question 20 ,bottom=5mm, boxrule=0.5mm, arc=2mm, outer arc=2mm]
<think>\\
To translate the natural language question into an executable SQLite query, let's break it down step by step:\\


\#\#\# Step 1: Understand the Question\\
The question asks: **"How many schools in Amador which the Low Grade is 9 and the High Grade is 12?"**\\

- **Key points**:\\
  1. We need to count the number of schools (`COUNT(*)`).\\
  2. The criteria for the schools are:\\
    - The `County Name` should be "Amador".\\
    - The `Low Grade` should be "9".\\
    - The `High Grade` should be "12".\\


\#\#\# Step 2: Identify the Relevant Table\\
From the schema, the `frpm` table contains the columns:\\
- `County Name`: To filter schools by county.\\
- `Low Grade`: To filter schools by the low grade.\\
- `High Grade`: To filter schools by the high grade.\\

This table is the primary focus of the query because:\\
- `County Name` directly corresponds to the county of interest ("Amador").\\
- `Low Grade` and `High Grade` are used to filter by grade levels.\\


\#\#\# Step 3: Apply the Filters\\
The query needs to:\\
1. Filter rows where `County Name = 'Amador'`.\\
2. Filter rows where `Low Grade = '9'`.\\
3. Filter rows where `High Grade = '12'`.\\

These filters ensure that only schools meeting all three conditions are counted.\\


\#\#\# Step 4: Count the Schools\\
Once the rows are filtered, the query should count the number of rows that satisfy the conditions. The `COUNT(*)` function is used for this purpose.\\


\#\#\# Step 5: Write the SQL Query\\
Based on the above reasoning, the SQL query is:\\
</think>\\
<answer>
SELECT COUNT(*) FROM frpm WHERE County Name = 'Amador' AND `Low Grade` = '9' AND `High Grade` = '12';\\
</answer>\\
\end{tcolorbox}

\begin{tcolorbox}[colback=mainboxbg, colframe=mainboxborder, coltitle=black, fonttitle=\bfseries, title=SLM-SQL-1.5B after CSC output results for question 30 ,bottom=10mm, boxrule=0.5mm, arc=2mm, outer arc=2mm]
<think>\\
To answer this question, we need to find the cities with the top 5 lowest enrollment numbers for students in grades 1 through 12. Let's analyze the given draft SQLs:\\

1. **First SQL Query**:\\
   ```sql\\
   SELECT      s.City FROM      frpm AS f INNER JOIN      schools AS s ON      f.CDSCode = s.CDSCode ORDER BY      f.`Enrollment (Ages 5-17)` ASC LIMIT 5\\
   ```\\
   - This query joins the `frpm` and `schools` tables based on the `CDSCode`.\\
   - It orders the results by the enrollment number in ascending order (`ASC`).\\
   - The `LIMIT 5` clause restricts the result to the top 5 cities.\\

2. **Second SQL Query**:\\
   ```sql\\
   SELECT      s.City FROM      schools s JOIN      frpm f ON      s.CDSCode = f.CDSCode GROUP BY      s.City ORDER BY      MAX(f.`Enrollment (Ages 5-17)`) ASC LIMIT 5\\
   ```\\
   - This query also joins the `frpm` and `schools` tables based on the `CDSCode`.\\
   - It groups the results by city.\\
   - It orders the results by the maximum enrollment number for each city in ascending order (`ASC`).\\
   - The `LIMIT 5` clause restricts the result to the top 5 cities.\\

Both queries are designed to find the top 5 cities with the lowest enrollment numbers for students. However, they approach the problem differently:\\
- The first query is more straightforward and directly orders by enrollment.\\
- The second query groups by city and then orders by the maximum enrollment, which is less intuitive for this specific question.\\

Given that the question asks for the cities with the top 5 lowest enrollment numbers, the first query is the correct one to use. It directly addresses the requirement by ordering the cities by their enrollment numbers.\\

<executionresult>\\
\[('Fremont',), ('Oakland',), ('Oakland',), ('Willows',), ('Long Beach',)\]\\
</executionresult>\\
</think>\\
<answer>\\
SELECT      s.City FROM      frpm AS f INNER JOIN      schools AS s ON      f.CDSCode = s.CDSCode ORDER BY      f.`Enrollment (Ages 5-17)` ASC LIMIT 5\\
</answer>\\
\end{tcolorbox}

\section{Prompt Templates}
\label{sec:appendix_prompt}

\onecolumn
\subsection{Prompt for SQL Generation}
\label{sec:prompt_candidate_sql_generate}

The SQL generation prompt adopts the prompt proposed by CSC-SQL\citep{sheng2025cscsqlcorrectiveselfconsistencytexttosql}.

\begin{tcolorbox}[colback=mainboxbg, colframe=mainboxborder, coltitle=black, fonttitle=\bfseries, title=SQL Generation Prompt,bottom=30mm, boxrule=0.5mm, arc=2mm, outer arc=2mm]
    You first thinks about the reasoning process in the mind and then provides the user with the answer.\\

    Task Overview:\\
    You are a data science expert. Below, you are provided with a database schema and a natural language question. Your task is to understand the schema and generate a valid SQL query to answer the question.\\
    
    Database Engine:\\
    SQLite\\
    
    Database Schema:\\
    \{DATABASE SCHEMA\}\\
    \\
    This schema describes the database's structure, including tables, columns, primary keys, foreign keys, and any relevant relationships or constraints.\\

    Question:\\
    \{EVIDENCE\}\\
    \{QUESTION\}\\
    
    Instructions:\\
    \mbox{-} Make sure you only output the information that is asked in the question. If the question asks for a specific column, make sure to only include that column in the SELECT clause, nothing more.\\
    \mbox{-} The generated query should return all of the information asked in the question without any missing or extra information.\\
    \mbox{-} Before generating the final SQL query, please think through the steps of how to write the query.\\
    
    Output Format:\\
    Show your work in <think> </think> tags. And return the final SQLite SQL query that starts with keyword `SELECT` in <answer> </answer> tags, for example <answer>SELECT AVG(rating\_score) FROM movies</answer>.  \\
    
    Let me solve this step by step.\\
\end{tcolorbox}

\subsection{Prompt for SQL Merge Revision}
\label{sec:prompt_sql_merge_revision}

The SQL merge revision prompt is slightly adjusted based on the CSC-SQL\citep{sheng2025cscsqlcorrectiveselfconsistencytexttosql} merge revision prompt. As shown in the red part below, we let the model analyze each draft SQL first, compare their differences, and finally generate the final SQL. At the same time, we emphasize that one of the two draft SQLs is correct, guiding the model to make the final decision.

\begin{tcolorbox}[colback=mainboxbg, colframe=mainboxborder, coltitle=black, fonttitle=\bfseries, title=SQL Merge Revision Prompt,bottom=10mm, boxrule=0.5mm, arc=2mm, outer arc=2mm]

    You first thinks about the reasoning process in the mind and then provides the user with the answer.\\

    Task Overview:\\
    You are a data science expert. Below, you are provided with a database schema, a natural language question, some draft SQL and its corresponding execution result. Your task is to understand the schema and generate a valid SQL query to answer the question.\\

    Database Engine:\\
    SQLite\\

    Database Schema:\\
    \{DATABASE SCHEMA\}\\
    \\
    This schema describes the database's structure, including tables, columns, primary keys, foreign keys, and any relevant relationships or constraints.\\

    Question:\\
    \{EVIDENCE\}\\
    \{QUESTION\}\\

    Here are some corresponding draft SQL and execute result:\\
    1. \{PREDICT\_SQL1\} \\
    Execution result\\
    \{EXECUTE\_RESULT1\}\\

    2. \{PREDICT\_SQL2\} \\
    Execution result\\
    \{EXECUTE\_RESULT2\}\\

    Instructions:\\
    \mbox{-} \textcolor{red}{You should first carefully analyze each draft SQL, compare their differences, and then conduct further analysis based on user questions to determine which draft SQL is correct in the end.}\\
    \mbox{-} \textcolor{red}{Remember that one of the draft SQLs is correct. You do not need to generate a new SQL combining their characteristics. Instead, output the draft SQL that you think is correct after careful analysis.}\\
    \mbox{-} Before generating the final SQL query, please think through the steps of how to write the query.\\
    
    Output Format:\\
    Show your work in <think> </think> tags. And return the final SQLite SQL query that starts with keyword `SELECT` in <answer> </answer> tags, for example <answer>SELECT AVG(rating\_score) FROM movies</answer>. \\ 
    
    Let me solve this step by step.\\
\end{tcolorbox}

\end{document}